\def\year{2019}\relax
\documentclass[letterpaper]{article} 
\usepackage{aaai19}  
\usepackage{times}  
\usepackage{helvet} 
\usepackage{courier}  
\usepackage[hyphens]{url}  
\usepackage{graphicx} 
\def\UrlFont{\rm}  
\usepackage{graphicx}  
\title{Integrating Automated Play in Level Co-Creation}
\author{Andrew Hoyt$^1$, Matthew Guzdial$^2$, Yalini Kumar$^1$, Gillian Smith$^3$, and Mark O. Riedl$^1$\\
$^1$School of Interactive Computing, Georgia Institute of Technology\\
$^2$Department of Computing Science, University of Alberta\\
$^3$Computer Science Department, Worcester Polytechnic Institute\\
andrewhoyt@gatech.edu, guzdial@ualberta.ca, ysk3@gatech.edu, gmsmith@wpi.edu, riedl@cc.gatech.edu
}

\begin{document}

\maketitle

\begin{abstract}
In level co-creation an AI and human work together to create a video game level. 
One open challenge in level co-creation is how to empower human users to ensure particular qualities of the final level, such as challenge. 
There has been significant prior research into automated pathing and automated playtesting for video game levels, but not in how to incorporate these into tools.
In this demonstration we present an improvement of the \textit{Morai Maker} mixed-initiative level editor for Super Mario Bros. that includes automated pathing and challenge approximation features. 
\end{abstract}

\section{Introduction}

\textit{Morai Maker} is a general video game level editor \cite{guzdial2017general}, which can load any set of sprites (game entities), which can then be used to construct new levels. 
However, it's development has focused on level design for the game Super Mario bros..
The editor has been augmented with a variety of machine learning-based AI level design partners who work with the human user in a turn-based fashion \cite{guzdial2018co}.
It has shown success in human subject studies with both novice and expert game designers \cite{guzdial2019friend}.

We identify two concerns raised by human users of \textit{Morai Maker}, based on the qualitative and quantitative results of prior human subject studies \cite{guzdial2019friend}.
First, that the AI partner can make additions to the level that are not guaranteed to be playable.
This concern arises from the fact that none of the four AI agents built for \textit{Morai Maker} has an explicit model of the physics in Super Mario Bros..
Thus there's no way for the AI partner to ensure that the additions to the level it creates allow for a fully playable level--a level that can be completed from beginning to end. 
However, this is not as simple as adding a playability check to any AI additions. 
Level design in \textit{Morai Maker} is intentionally iterative.
Therefore at any point in time the level may be unplayable because it is not yet complete.

The second concern is that there's no way in the editor to ensure level content is reachable without the user playing the level.
While \textit{Morai Maker} allows users to play through the current level, this can be challenging and frustrating given that it can break up the flow of level design.
It also forces level creators to only design levels that they personally can play themselves, if they care about level playability.

To address these two concerns in this demonstration paper we present two new features for \textit{Morai Maker}. 
The first is an A* Reachability Check, which allows a user to determine whether it is possible as Mario to go from any two arbitrary points. 
The second is an A*-based Survival Analysis \cite{isaksen2015exploring}, giving users a rough approximation of the difficulty of navigating some section of the level.
We briefly discuss related, prior work and then detail these two new features.

\section{Related Work}

\textit{Morai Maker} can be considered a co-creative example of procedural level generation via machine learning (PCGML) \cite{summerville2018procedural}.
Such tools have been previously discussed in the literature \cite{summerville2018procedural,zhuexplainable}, but are still underexplored. 
Comparatively there exist many prior approaches to co-creative or mixed-initiative level design without machine learning \cite{yannakakis2014mixed,deterding2017mixed}. 
These approaches instead rely upon search or grammar-based approaches \cite{smith2011tanagara,liapis2013sentient,shaker2013ropossum,baldwin2017mixed,alvarez2018fostering}.
However, none of these prior approaches include automated playtesting and challenge analysis systems like the features presented in this paper.

Cicero \cite{machado2018ai} is a mixed-initiative game development tool built within the General Video Game AI (GVGAI) framework \cite{perez2018general}. 
Therefore, it also has access to a variety of general GVGAI AI agents for automated playtesting \cite{horn2016mcts,khalifa2016modifying}. 
However, because these agents are general they cannot be employed in the same way as the features we present, which rely on finding the optimal path.

Automated play has a rich history in the field of game AI. 
Our work specifically relates to the Mario AI Benchmark, Competitions, and Championships \cite{karakovskiy2012mario,togelius2013mario}. 
Specifically, the discovery that an A* agent could outperform all other automated game playing agents for Super Mario Bros. when given access to the game's rules. 
This directly lead to our choice to rely on A* as the basis for both new features.
We draw on Isaksen et al.'s conception of survival analysis \cite{isaksen2015exploring,isaksen2015comparing} as an approach for determining game difficulty by simulating many agents through a game or level.

\begin{figure}[tb]
\centering
	\includegraphics[width=3in]{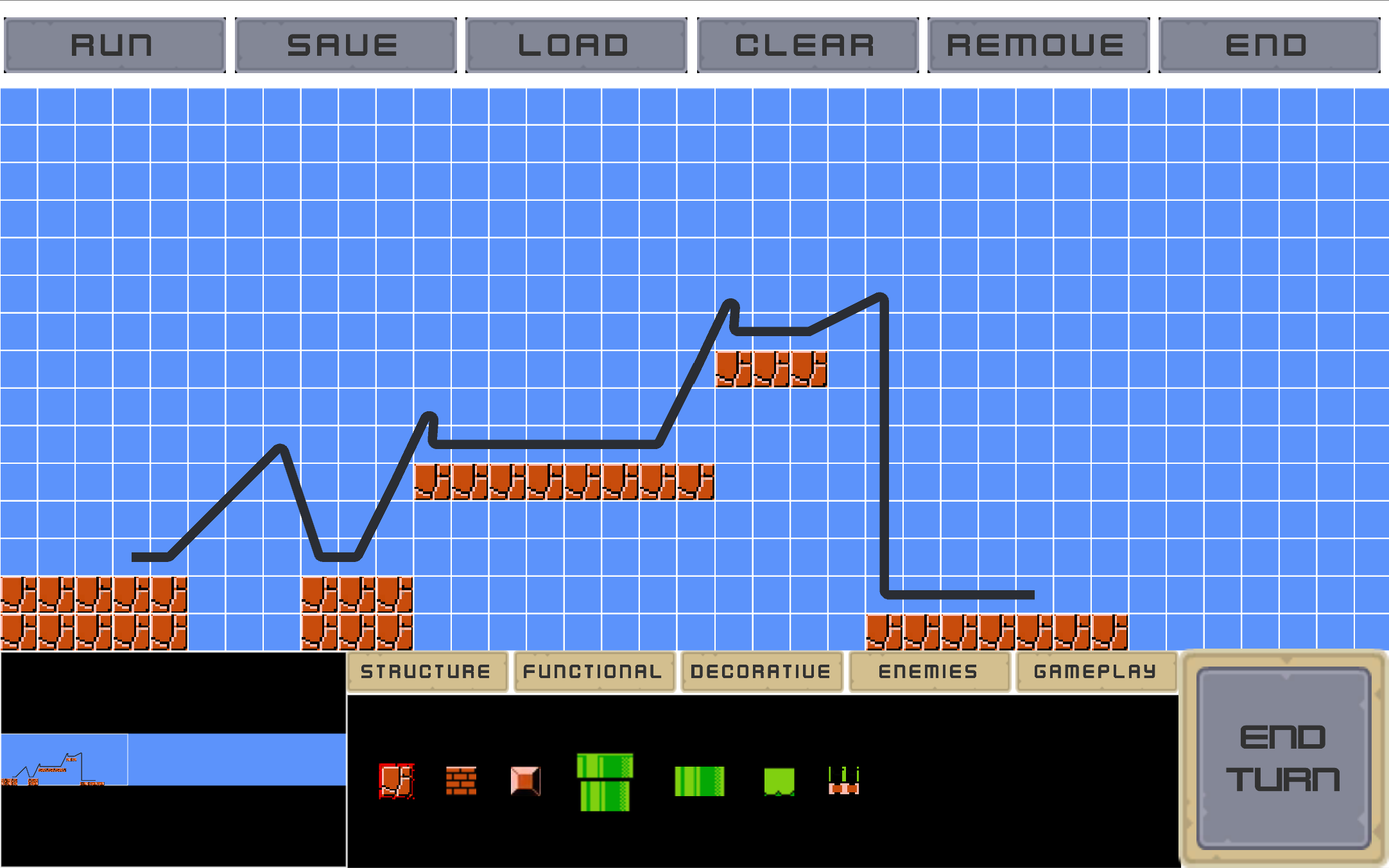}
	\caption{Screenshot of a calculated path displayed over a user-constructed level section.}
	\label{fig:pathinScreenshot}
\end{figure}

\section{A* Reachability Check}

The A* Reachability Check is the first of the features we introduce to address user concerns with level playability and challenge. 
We visualize an example output of the feature in Figure \ref{fig:pathinScreenshot}. 
You can see some example level geography created by a user of the system (the orange ground blocks). 
To employ the reachability check a user pressed the ``P'' key on the keyboard and then clicks somewhere with the mouse. 
A black line is drawn from where the first mouse click occurs until the mouse is clicked again.
At this point, if it is possible given the mechanics of Super Mario Bros. for a player to traverse from the first point to the second the optimal path is drawn.
Otherwise no line is drawn.
This feature required the representation of Super Mario Bros. physics as A* path finding operators \cite{karakovskiy2012mario,togelius2013mario}.

This feature does not guarantee that the visualized path is the only possible path, but it does allow a user to check if any path is possible between two points. 
Thus users can check whether a level is possible to complete without having to or even having to be able to complete the level themselves.
However, this feature alone does not reflect the potential difficulty of paths outside of the one visualized path.

\begin{figure}[tb]
\centering
	\includegraphics[width=3in]{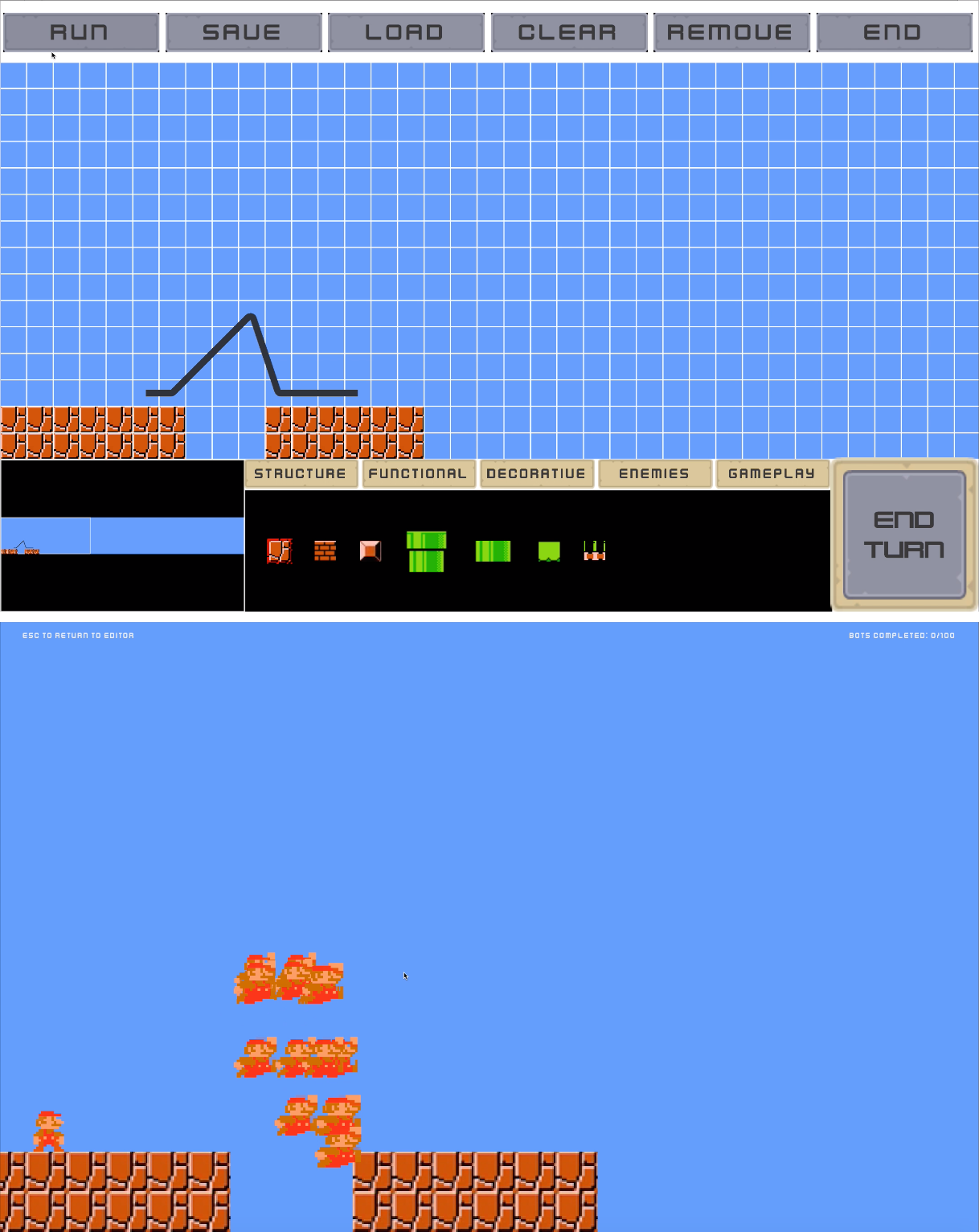}
	\caption{Example of the path (top) and corresponding in-progress survival analysis (bottom).}
	\label{fig:survivalAnalysis}
\end{figure}

\section{A*-based Survival Analysis}

The second feature is based upon Isaksen et al.'s Survival Analysis, in which many agents attempt to play through a level \cite{isaksen2015exploring,isaksen2015comparing}. 
These agents differ in the optimality of their play, which means one can employs Survival Analysis to determine a rough notion of challenge in a level (e.g. how un-optimal can one play and still make it to the end). 
Our second feature presents a version of this for Super Mario Bros., which takes the A* path and sends a designer-chosen number of agents along this path (with a default of 100).
Each agent has a randomly chosen threshold of how close they must be to a particular node on the A* path before attempting to reach the next one. 
This allows for the variance in optimality of these agents. 
To interact with the feature a user makes a path via the reachability check feature as seen on the top of Figure \ref{fig:survivalAnalysis}.
If there is a current path, hitting run will send the agents along that path, as can be seen at the bottom of Figure \ref{fig:survivalAnalysis}.
The top right corner of the screen will then count up the number of agents that make it to the final goal, as a rough approximation of challenge. 

Due to the variability in assigning random thresholds to A* agents, the final proportion of successful agents can differ across new runs of the same path. 
In addition, this proportion should not be taken as equivalent to perceived human difficulty, as humans play differently than A* agents. 
However, it can still be a useful measure for comparing difficulty of different level sections built within the editor.

\section{Acknowledgments}

This material is based upon work supported by the National Science Foundation under Grant No. IIS-1525967. Any opinions, findings, and conclusions or recommendations expressed in this material are those of the author(s) and do not necessarily reflect the views of the National Science Foundation.

\bibliographystyle{aiide}
\bibliography{aiide}
\end{document}